\documentclass[sn-nature]{sn-jnl}% Style for submissions to Nature Portfolio journals

\AtBeginDocument{%  
  \newgeometry{margin=1in} % or any margin you prefer  
}

\usepackage{graphicx}%
\usepackage{multirow}%
\usepackage{amsmath,amssymb,amsfonts}%
\usepackage{amsthm}%
\usepackage{mathrsfs}%
\usepackage[title]{appendix}%
\usepackage{xcolor}%
\usepackage{textcomp}%
\usepackage{manyfoot}%
\usepackage{booktabs}%
\usepackage{algpseudocode}%
\usepackage{listings}%
\usepackage{longtable}%
\usepackage[most]{tcolorbox}%
\usepackage{booktabs} % For prettier tables
\usepackage{tabularx}
\usepackage{float}
\usepackage{courier}
\usepackage[linesnumbered,ruled,vlined]{algorithm2e}
\usepackage{hyperref}  
\usepackage{xurl}  
\SetKwFor{ForPar}{for (in parallel)}{do}{endfor}  

\tcbset{
  aibox/.style={
    width=474.18663pt,
    top=10pt,
    colback=white,
    colframe=black,
    colbacktitle=black,
    enhanced,
    center,
    attach boxed title to top left={yshift=-0.1in,xshift=0.15in},
    boxed title style={boxrule=0pt,colframe=white,},
  }
}
\newtcolorbox{AIbox}[2][]{aibox,title=#2,#1}

\begin{document}

\title[Article Title]{Cognitive Loop via In-Situ Optimization: Self-Adaptive Reasoning for Science}

\author[]{\fnm{Newman} \sur{Cheng}\footnote{Equal Contribution\\Correspondence: \texttt{\{newmancheng, b-gbroadbent, willchap\}@microsoft.com}}}
\author[]{\fnm{Gordon} \sur{Broadbent}\footnotemark[1]}  
\author[]{\fnm{William} \sur{Chappell}}

\affil{Microsoft Discovery and Quantum, Office of the Chief Technology Officer}

\abstract{The capacity for artificial intelligence (AI) to formulate, evolve, and test altered thought patterns under dynamic conditions indicates advanced cognition that is crucial for scientific discovery. The existing AI development landscape falls into two categories: 1) frameworks over non-reasoning models that natively incorporate opinions on how humans think, and 2) reasoning models that abstract precise control of the reasoning intuition away from end users. While powerful, for scientists to maximize utility of AI in scientific discovery, they not only require accuracy and transparency in reasoning, but also steerability. Hence, we introduce an alternative approach that enables deep and precise control over the reasoning process called: a cognitive loop via in-situ optimization (CLIO). CLIO enables large language models (LLMs) to self-formulate ways of approaching a problem, adapt behavior when self-confidence is low, and ultimately provide scientists with a final belief or answer. Through CLIO’s open design, scientists can observe uncertainty levels, understand how final belief states are formulated using graph structures, and interject corrections. Without any further post-training, OpenAI's GPT-4.1 with CLIO yields an accuracy of 22.37\% in text-based biology and medicine questions on Humanity’s Last Exam (HLE). This yields a 13.82\% net or 161.64\% relative increase when compared to the base GPT-4.1 model and surpasses OpenAI's o3 performance in high and low reasoning effort modes. We further discovered that oscillations within internal uncertainty measures are key in determining the accuracy of CLIO’s results, revealing how its open design and internal mechanisms can provide insight and control into scientific decision-making processes.}

\maketitle
\section{Introduction}
Long-running LLM agents that reason, plan, and execute high-value tasks over long temporal periods, such as in the creation of new materials or drugs, are poised to transform scientific discovery. Currently, methods to produce long-form reasoning are based on post-training cycles prior to model deployment. These corresponding models possess reasoning strategies and instinctual properties dictated by model providers without direct input from end users, relegating them to steer behavior utilizing the context window alone. In this paper, we demonstrate an alternative approach that enables end users to steer and correct thought patterns in real-time or post-hoc while achieving the same or greater-level intelligence when compared to a post-trained model. Our approach, CLIO, is designed as an alternative or complement to the reinforcement learning post-training step with recursive reflection to increase non-reasoning models’ ability to think through problems and select the best approach.\\

\noindent While HLE scores are increasing from new models, with trends showing that it will be saturated akin to other benchmarks, accuracy is no longer the only measure of success. Now, the real challenge is active inclusion of the end user, bypassing passive user observation. To address this challenge, we tested CLIO's ability to answer biology and medicine questions from HLE with no tools. A model-only approach allows for a direct test of the efficacy between reinforcement learning models when compared to in-situ optimization of non-reasoning models. CLIO does more than just increase performance; CLIO brings the end user along throughout the creation of the entire thought process, increasing performance and explainability into internal behavior. Metrics, such as uncertainty oscillation or number of new brainstorming perspectives, enable an understanding of when results and decisions being made by the model can be trusted. As we embark on this new age of discovery where answers will be as good or better than human subject matter experts, the challenge has shifted to creating the balance between human and machine teaming. Understanding when experts need to interject without oversaturation and ensuring critical decisions are steerable is crucial. Ultimately, CLIO puts these principles at the forefront. 

\section{Prior Work}
\subsection{Reasoning Approaches}
The neuroplasticity of the human brain is central to how humans learn, remember, and adapt.\cite{GAZERANI2025149643} As such, human brains possess the ability to create, modify, or remove neural connections based on experiences.\cite{PASCUALLEONE2006315} Designing cognitive-like systems that can think increasingly like humans requires the fundamental ability to formulate different approaches to problems, receive external feedback, and, subsequently, pivot strategies.\cite{Qadir2025, Don-Yehiya2025, Kasdin2025, Mohar2025} Searching for new strategies when unsuccessful and saving useful patterns for reuse is key in scaling developed knowledge.\cite{Clewett2019, guo2025deepseek} This approach is central to post-training models today and is the cornerstone of reinforcement learning (RL). Evidence shows continual scaling as reasoning models solve increasingly complex, multi-step reasoning problems within verifiable domains.\cite{wen2025reinforcement, zhao2025learning} \\

\noindent The current phase of RL training predominantly utilizes externally verifiable rewards (RLVR)to provide feedback to the model.\cite{su2025crossing} An example of RLVR is a binary reward for whether an LLM-produced code segment compiled or not. Recent works, like “Learning to Reason without External Rewards,” focus on the use of RL training with internal feedback (RLIR), which trades externally verifiable and human-defined rewards in favor of internal confidence signals or self-certainty.\cite{zhao2025learning} RLIR results demonstrate how internally computed or estimated rewards can scale performance during post-training. In addition, RLIR aims to reduce dependence solely on externally verifiable human rewards utilized in reinforcement learning with human feedback (RLHF). However, end users utilizing RLVR, RLIR, or RLHF post-trained models all share the same fundamental limitations: the learning or reasoning process elicited is neither personal, nor steerable as learning terminates once the trained model is handed to the user.\cite{korbak2025chain} Enabling such customization requires further post-training on high-quality data which often does not exist for cutting-edge scientific discoveries.\cite{xu2025towards}

\subsection{Systems that Evolve Knowledge}
Recent work by Google’s AlphaEvolve demonstrates how AI systems can continue to scale inference-time compute to discover or find known state-of-the-art algorithms through an evolutionary process.\cite{novikov2025alphaevolve} Throughout AlphaEvolve’s discovery process, the use of verifiable rewards ranks and prioritizes subsequent algorithms that AlphaEvolve should use to evolve, reducing the burden on LLMs alone to control the direction of exploration. This prioritization is critical as it narrows the search space and complexity necessary for downstream evolution. Hence, it enables the incremental evolution of knowledge, that is lower in complexity, to be aggregated and used in answering highly complex problems. As Apple’s “Illusion of Thinking” paper highlights, models are particularly performant in solving problems that contain a smaller combinatorial or exponential search space, but quickly fall apart as that space grows linearly or exponentially.\cite{shojaee2025illusion} By reducing the complexity of problems required for LLMs to address at each individualized state, complex search spaces can theoretically be mapped and reduced while considering multiple signals of human feedback.\cite{SLADKY2024223, SHEA2014186} Low complexity spaces can be explored using breadth-wise approaches while high complexity spaces require depth wise approaches. Breadth-wise exploration requires a greater sampling of different perspectives or paths.\cite{doi:10.1126/sciadv.adn5290, chowdhery2023palm} Depth-wise approaches require consecutively building out reasoning paths, prioritizing the most promising ones, and pruning those that are least likely to yield success.\cite{Silver2016} 

\subsection{Science Agents}
Google’s Co-Scientist and Sakana’s AI Scientist have identified the potential utility for long-running agents to assist in scientific discovery.\cite{gottweis2025towards, lu2024ai} More recent works like TxGemma: Efficient and Agentic LLMs for Therapeutics make use of the HLE benchmark to evaluate their system.\cite{wang2025txgemma} Similarly, Stanford’s BioMNI: General-Purpose Biomedical AI Agent demonstrates how the use of combining an agentic system with well-curated scientific literature and scientific tools can improve performance on a subset of HLE biology and medicine questions.\cite{Huang2025.05.30.656746} Both focus on the demonstration of their respective approaches in boosting performance, rather than critical areas of control, central to end users. In practice, FutureHouse’s work with Robin utilized a multi-agent architecture to identify a novel treatment for dry age-related macular degeneration and paired it with an iterative lab-in-the-loop to automate portions of the scientific discovery process.\cite{ghareeb2025robin} This has been extended further where “self-driving laboratories” are being used to accelerate the discovery of new materials.\cite{Delgado-Licona2025} While powerful, each of these approaches demonstrates areas of concern when scientists are unable to control the reasoning process of LLMs. The lack of steerability in deeply technical and high-stakes domains like scientific discovery or diagnostic reasoning is a clear gap.\cite{gridach2025agentic} We address these limitations through CLIO.

\section{Methods}
Our approach leverages the inference-time evaluation of progress and self-confidence to optimize thinking in real time without additional training cycles. Furthermore, unlike the post-training reinforcement learning process which requires high-quality data, CLIO does not require additional data or training. Instead, our approach makes use of internal self-confidence and reflection to reason through and adapt to problems at inference time. While the need for additional training data or compute cycles is obviated, the core challenge comes in a practical system design that prioritizes the shape of human cognitive behaviors from the outset.\footnote{\textit{n. the process by which knowledge and understanding is developed in the mind}}    

\subsection{The Recursive Nature of Cognition}
The human thought process is not always linear, nor is it direct.\cite{Bieberich2012} For example, problem solving is non-linear in nature as it often requires both the formulation of the potential paths that may emerge and selection of the most promising paths.\cite{ferrigno2020recursive} Through this framing, we designed CLIO to possess both breadth- and depth-wise exploration capabilities. To enable breadth, CLIO builds on the inspiration of existing approaches like chain-of-thought prompting, and reasoning then acting (ReAct) when using tools.\cite{wei2022chain, yao2023react} The key difference from other systems is that CLIO does not rely on prompting to construct thought patterns, rather orchestration of thought is native to its design. This orchestration pattern still follows the “feed-forward” nature of knowledge, allowing CLIO to explore many different options as desired.\\

\noindent Conversely, to enable depth, CLIO introduces a new utilization of recursion that balances the challenge of controlling semantic explosion in adaptive contexts with algorithmically defined constraints. CLIO’s ability to invoke itself enables the formulation of individualized-thought channels that possess independent context windows. These clean context windows are crucial when CLIO dives into a particular area of exploration, without polluting the aggregated context with incomplete thoughts. Hence, this design enables exploration across many different avenues in depth and adaptation to changes, when necessary, within or across thought channels.\\

\noindent To control CLIO’s depth of exploration, we enabled the system to self-reflect and determine the necessary duration for how long it needs to self-reflect based on a maximum recursion, or maximum cognitive depth. We observed that without the proper algorithmic control on recursion, these systems continue to goad themselves into further exploration, leading to fundamentally incorrect results over long periods of time. Hence, control over cognitive depth is crucial and akin to reasoning effort levels from models like o3, where low/medium/high effort levels are configurable. The key difference is that CLIO controls both the depth of exploration and the iterative assumptions made at each step. Each of these assumptions can be altered or weighed differently. While exploring, CLIO can self-optimize its own internal strategy through a series of parameters that are editable when invoking itself. These areas enable CLIO’s persona to be modified, changing the area of focus or problem-solving approach. Most commonly, we observe CLIO utilizing this optimization component to resolve uncertainties it has self-recognized throughout the execution process. Hence, this utilization at runtime directly influences the optimization of the system’s internal belief structures over time.\\
\begin{figure*}
\centering  
\begin{minipage}{0.92\textwidth}  
\begin{algorithm}[H]  
\caption{CLIO's Recursive Algorithm}  
\KwIn{Initial semantic state $s_0=\phi(\text{question OR prompt})$; branching factor $b$; maximum depth $D$; confidence threshold $\tau$; current depth $d$; confidence $c$}  
\KwOut{Synthesized answer from LLM based on terminal/high-confidence states}  
\textbf{Definitions:}\\  
- $T(s)$: Completion (terminal) Function; returns $0$ if $s$ is a solved/terminal state, otherwise $>0$.\\  
- $c(s)$: Confidence Function; returns a real-valued confidence score for state $s$.\\  
- $\pi(\cdot|s)$: LLM samples possible next states from $s$.\\  
- $\text{LLM}(\mathcal{S}, s)$: LLM synthesizes answer from $\mathcal{S}$ and current context $s$.\\  
  
\SetKwFunction{FCLIO}{CLIO}  
\SetKwProg{Fn}{Function}{:}{}  
\Fn{\FCLIO{$s,\,d$}}{  
    \If{$T(s)=0$ \textbf{or} $c(s)\ge \tau$}{  
        \Return $\{s\}$  
    }  
    \If{$d \ge D$}{  
        $\mathcal{S} \leftarrow \emptyset$\;  
        $i \leftarrow 0$\;  
        \While{$i < b$}{  
            $s' \leftarrow \pi(\cdot|s)$  
            \If{$T(s')=0$ \textbf{or} $c(s')\ge \tau$}{  
                $\mathcal{S} \leftarrow \mathcal{S} \cup \{s'\}$\;  
                \textbf{break}  
            }  
            $i \leftarrow i+1$\;  
        }  
        \Return $\mathcal{S}$  
    }  
    $\{s_1, s_2, \ldots, s_b\} \leftarrow \pi(\cdot|s)$\;  
    $\mathcal{S} \leftarrow \emptyset$\;  
    \ForPar{$i \leftarrow 1$ \KwTo $b$}{  
        $\mathcal{S} \leftarrow \mathcal{S} \cup$ \FCLIO{$s_i,\,d+1$}  
    }  
    \Return \textbf{$\text{LLM}(\mathcal{S}, s)$}  
}  
\end{algorithm}  
\end{minipage}  
\label{alg:clio-recursive-algorithm}  
\caption{CLIO’s recursive nature with semantic stopping through self-determination of completion}
\end{figure*}

\subsection{Building Awareness into CLIO}
Within each thought channel, CLIO’s runtime is aware of when to complete its thought processes. To do so, we enabled a special “completion” function that can be invoked to terminate the thought channel. However, proper invocation of this termination function requires both sufficient confidence and a temporal understanding of the progress made by the system. Integrating temporal awareness is key, as the minimization of token-level probabilities alone fails to account for the notion of time. Hence, to prevent early termination, we enabled CLIO to dynamically register the special completion function only when a certain threshold of coverage was met. Under this construct, reduction of bias is critical when solving complex challenge problems that possess potentially multiple correct answers. The precedence of those assumptions contains minimal variations in themselves but leads to highly influential effects downstream when reasoning through and selecting the proper approach. Across these possible reasoning paths constructed during the breadth- and depth-wise exploration phase, the mechanism of attention serves as a critical system internal process. Thereby, creating a belief state regarding the significance of various elements can be established. Conceptually, this process is akin to human behavior where humans adopt diverse cognitive perspectives or reasoning strategies before formulating an opinion on plausibility or preference.\cite{AdahMiller2025}

\subsection{Overcoming the Over-Indexing Challenge}
Agentic optimization approaches often ensemble multiple different perspectives to enable diversity of thought and sampling, to elicit a correct answer. The most common approach to reduce and select an answer from the sampled variations has been a custom prompt-based, facing both limitations in the quality of the prompt as well as bias native within the context window itself.\cite{liu2023lost, hsieh2024ruler} To mitigate these issues raised through ensembling, we leveraged graph structures to reduce noise.\\

\noindent To build these graph structures, we utilized GPT-4.1 to perform entity and relationship extraction from CLIO’s thought processes. These extractions then underwent unsupervised clustering to form hierarchical communities of information that were subsequently summarized and embedded. At runtime, these summaries and embeddings were used to assist in the retrieval process for relevant pieces of information related to CLIO’s query.\cite{edge2024local} To build a holistic view, we enabled a control knob for CLIO to think more, resulting in multiple different runs, often using different temperature configurations. Each run produced a distinct chain of thought that we subsequently collected and built a joint graph of, using all sampled sequences. To construct the final graph, we leveraged the use of LLM-induced graph structures with unsupervised clustering to produce a representation across all runs. Finally, CLIO produced a response by querying the final graph representation using the original question provided.

\setcounter{AlgoLine}{0}   %  ← resets the line numbering  

\begin{figure*}
\centering  
\begin{minipage}{0.95\textwidth}  
\begin{algorithm}[H]  
\caption{\textsc{CLIO's Ensembling and Reduction}}  
\KwIn{Semantic state $s$; depth $d$; max sampling $M$; folds $f$, follow\_up $u$}  
\KwOut{Final answer $\hat{a}$}  
\SetKwFunction{FCLIO}{CLIO}  
\SetKwFunction{FCLIOMore}{CLIO\_MoreThinking}  
\SetKwProg{Fn}{Function}{:}{}  
\Fn{\FCLIOMore{$s,\,d,\,M,\,f,\,u$}}{  
    $\mathcal{P} \leftarrow \emptyset$\;  
    \For{$i \leftarrow 1$ \KwTo $M$}{  
        $\pi_i \leftarrow$ \FCLIO{$s,\,d$}\;  
        $\mathcal{P} \leftarrow \mathcal{P} \cup \{\pi_i\}$\;  
    }  
    $G \leftarrow \textsc{BuildGraphFromPaths}(\mathcal{P})$\;  
    $\mathcal{C} \leftarrow \textsc{ClusterGraph}(G)$\;  
    \ForEach{community $c \in \mathcal{C}$}{  
        $summary_c \leftarrow \textsc{LLM\_Summarize}(G[c], s)$\;  
        \ForEach{node $v \in c$}{  
            $G[v].\text{summary} \leftarrow summary_c$\;  
        }  
    }  
    $\hat{a} \leftarrow \textsc{DRIFT\_SEARCH}(s, G, f, u)$\;  
    \Return $\hat{a}$\;  
}  
\end{algorithm}  
\end{minipage}  
\label{alg:cliomorethinking}  
\caption{CLIO’s utilization of a graph-induced structures to balance beliefs during increased thinking modes}
\end{figure*} 

\section{Evaluation}
\begin{figure}[htbp]  
    \centering  
    \includegraphics[width=\textwidth]{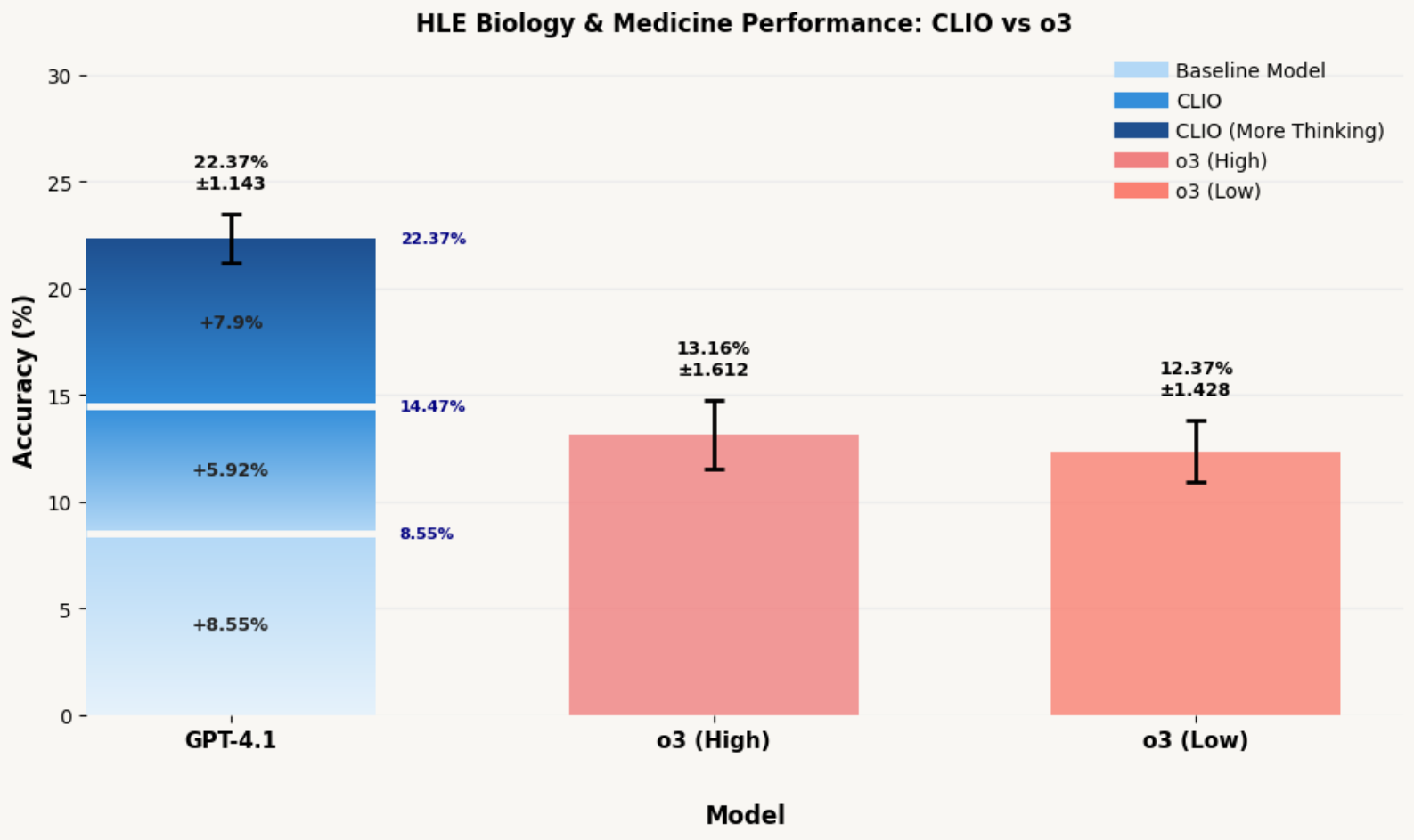}
    \caption{GPT-4.1 with CLIO demonstrates on par performance with o3 in high and low reasoning efforts, while CLIO with More Thinking enables a leap in performance.}  
    \label{fig:CLIO-accuracies}  
\end{figure}

To evaluate CLIO's performance against reasoning models, we compared our system without access to external tools or data against o3 in high- and low-reasoning effort modes. We extended our study to include GPT-4.1 and GPT-4o to illustrate how we can elevate the performance of completion models into those on par with reasoning-class models, all without post-training. Ultimately, we demonstrated how cognitive systems designed with the proper abstractions can be orchestrated to 1) self-formulate thought patterns, 2) self-formulate confidence and feedback, and 3) self-recognize success or failure. We further conducted a series of ablation studies that measured the system’s performance from multiple aspects: accuracy with more thinking, similarity of reasoning structure, belief reduction using graph structures, and uncertainty analysis. Overall, we found that CLIO surpassed o3 in both high- and low- reasoning effort modes on the 152 text-based Biology/Medicine questions from Humanity’s Last Exam in Pass @ 1 scenarios. On top of accuracy alone, we ran five independent Pass @ 1 scenarios to demonstrate CLIO’s stability and fundamentally less variability than that of o3. Figure \ref{fig:CLIO-accuracies} below denotes the performance and variability of GPT-4.1 with CLIO when compared to o3.\\

\noindent Furthermore, our results demonstrated how the utilization of graph structures reduced sporadic noise from stronger false positive arguments, leading to an increasingly balanced perspective when answering the question. This was evident through the performance increase of 7.90\% between base GPT-4.1 with CLIO and CLIO paired with “more thinking”. To obtain our Pass @ 1 metric, we leveraged dynamic reasoning and inference with flexible traversal (DRIFT) search on the ensembled graph representation. DRIFT search is a technique that combines global and local queries to obtain information spanning multiple conceptual representations in the graph.\cite{whiting2024driftsearch} Table \ref{tab:fullperformance} denotes our complete Pass @ 1 accuracies within the Biology/Medicine domain, with a detailed breakdown of specific question category performance within them.\\

\begin{table*}[hbp] 
\centering  
\begin{minipage}{\textwidth}  
\begin{tabularx}{\textwidth}{|l|X|X|X|X|}  
\hline  
\textbf{Question Category} & \textbf{o3 (High)} & \textbf{o3 (Low)} & \textbf{GPT-4.1} & \textbf{GPT-4.1 w/ CLIO} \\  
\hline  
Bio/Med - Genetics & 0/27 & 2/27 & 2/27 & 7/27 \\  
\hline  
Bio/Med - Biology & 3/31 & 1/31 & 0/31 & 8/31 \\  
\hline  
Bio/Med - Ecology & 5/20 & 2/20 & 3/20 & 7/20 \\  
\hline  
Bio/Med - Neuroscience & 3/17 & 5/17 & 2/17 & 4/17 \\  
\hline  
Bio/Med - Biochemistry & 3/16 & 1/16 & 1/16 & 2/16 \\  
\hline  
Bio/Med - Microbiology & 0/9 & 1/9 & 2/9 & 1/9 \\  
\hline  
Bio/Med - Immunology & 1/8 & 1/8 & 1/8 & 2/8 \\  
\hline  
Bio/Med - Molecular Biology & 1/8 & 0/8 & 0/8 & 1/8 \\  
\hline  
Bio/Med - Computational Biology & 0/7 & 0/7 & 0/7 & 1/7 \\  
\hline  
Bio/Med - Biophysics & 2/4 & 2/4 & 2/4 & 1/4 \\  
\hline  
Bio/Med - Bioinformatics & 2/3 & 1/3 & 0/3 & 1/3 \\  
\hline  
Bio/Med - Genomics & 0/1 & 0/1 & 0/1 & 0/1 \\  
\hline  
Bio/Med - Physiology & 1/1 & 1/1 & 0/1 & 0/1 \\  
\hline  
\textbf{TOTAL} & \textbf{21/152 = 13.81\%} & \textbf{17/152 = 11.18\%} & \textbf{13/152 = 8.55\%} & \textbf{34/152 = 22.37\%} \\  
\hline  
\end{tabularx}  
\end{minipage}  
\caption{Pass @ 1 Accuracy for text-based Biology/Medicine HLE Questions}  
\label{tab:fullperformance}  
\end{table*}

\begin{figure}[htbp]
    \centering  
    \includegraphics[width=\textwidth]{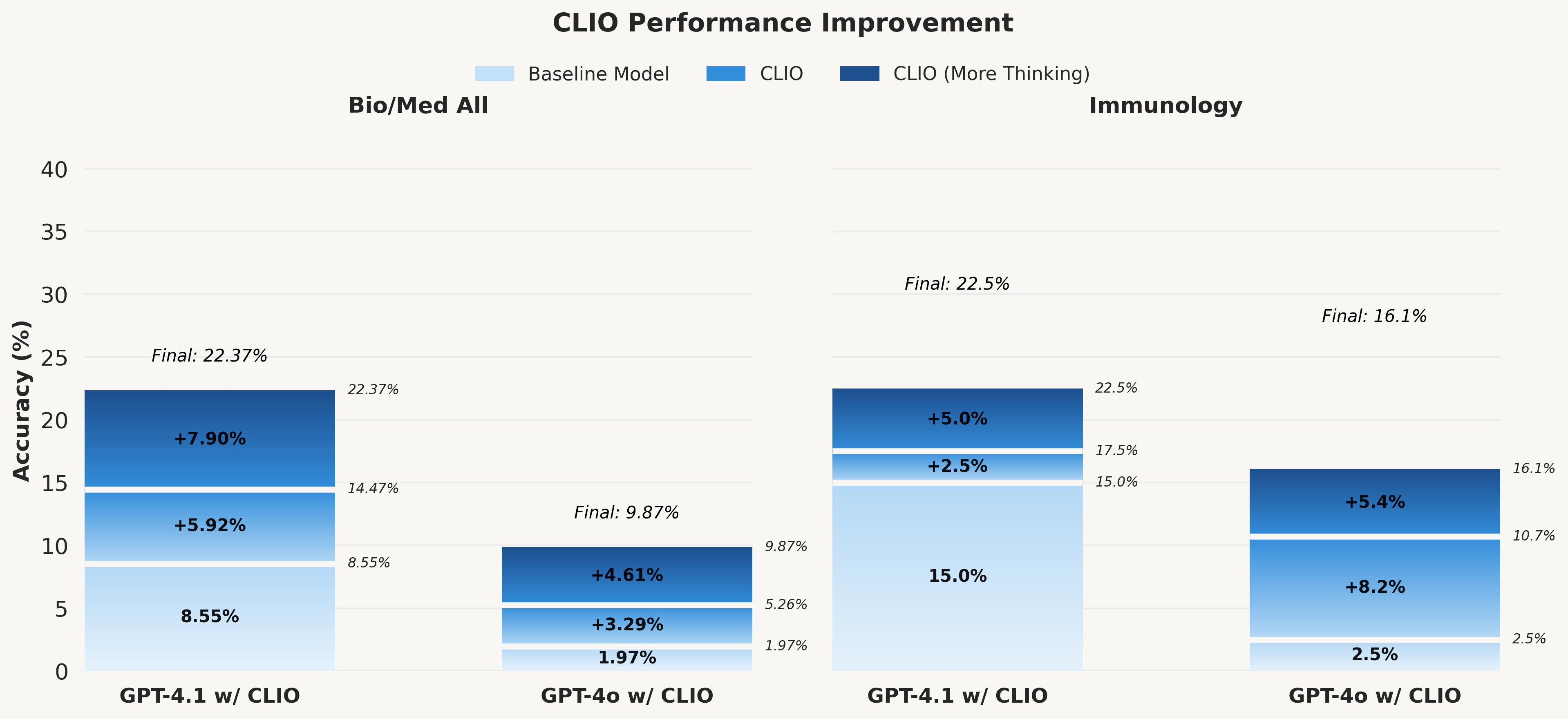}
    \caption{Increasing CLIO’s performance through increased thinking}  
    \label{fig:clio_performance_improvement}  
\end{figure}

\noindent Across multiple runs of CLIO, we found the graph-based structure alongside a multi-resolution query of information reduced variance and raised the overall stability of the system’s performance, even when multiple different perspectives were sampled. Figure \ref{fig:CLIO-accuracies} denotes the range in standard deviations between GPT-4.1 with CLIO, of which is smaller than o3. To further validate the potential of graph-induced structures and different search techniques to include DRIFT search, we queried the final graph constructed per question $n$ times. Specifically, we ran questions in the immunology category $n=5$ times for both GPT-4o, GPT-4.1, and CLIO backed by both models. In Figure \ref{fig:clio_performance_improvement}, CLIO's approach to more thinking did not require a random selection and yielded a 22.5\% average accuracy across all immunology questions in this case.

\section{Discussion}
\subsection{Trust of Thought}
\begin{figure}[htbp] 
    \centering  
    \includegraphics[width=\linewidth]{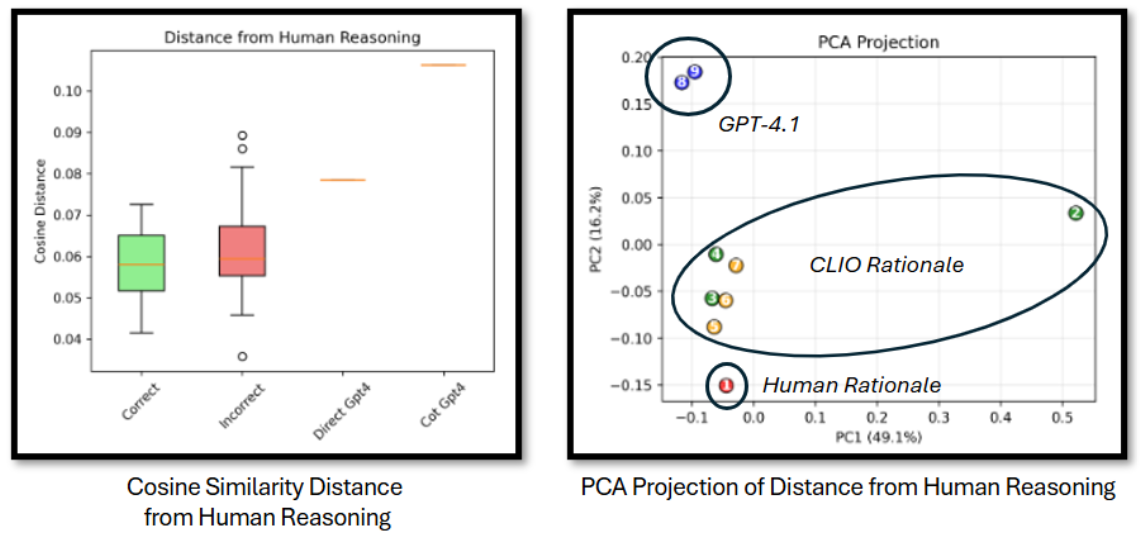}  
    \caption{CLIO’s produced chain of thoughts are more similar to human reasoning}  
    \label{fig:trust-of-thought}  
\end{figure}

We studied the diversity and semantic distance between all correct and incorrect chain-of-thought sequences produced by CLIO and plotted them against human-annotated rationales provided by Humanity’s Last Exam. For each of the 34 distinct questions that CLIO successfully answered, we performed additional sampling to ensure 95\% statistical power. We found that the rationales produced by CLIO are more like the rationales produced by humans when compared to the base model itself. Figure \ref{fig:trust-of-thought} denotes a particular example for an immunology question that both CLIO and GPT-4.1 can solve. Here, we plotted the similarity between CLIO’s and GPT-4.1’s chain of thought by projecting their embeddings, generated using the text-embedding-3-large model, onto a 2D space with the top two computed principal components. GPT-4.1 was tested using the exact HLE question as well as detailed chain-of-thought prompting to solicit clearer reasoning patterns that are ideally in line with human annotations. 

\subsection{Enabling Deep and Precise Reasoning Control}
CLIO's open nature of exposing its belief state makes it ripe for control. It is eminently important for a human to communicate their cognitive processes regarding the logic of “how” they achieved a certain result or the “why” behind a correct answer selection. Controlling and understanding this belief state requires explainability, traceability, and the ability to argue for or against cognitive processes. Our approach to designing CLIO puts this at the forefront, where we emphasized the ability to not only explain the thought process and reasoning for tracking and testing, but also to raise flags when there is uncertainty. A lack of such control greatly reduces value in deep applications, as reasoning without control or correction is non-personal and can lead to indefensibility in the scientific context. The current design of reasoning-class models inhibits the ability for humans to communicate and precisely correct a reasoning chain for imprecise or non-preferable paths.\cite{gridach2025agentic} Untrustworthy or uncontrollable systems are equivalent to accepting academic research that fails to meet peer review requirements.

\subsubsection{Immunology Case Study}
\begin{figure}[htbp]
    \centering
    \includegraphics[width=\linewidth]{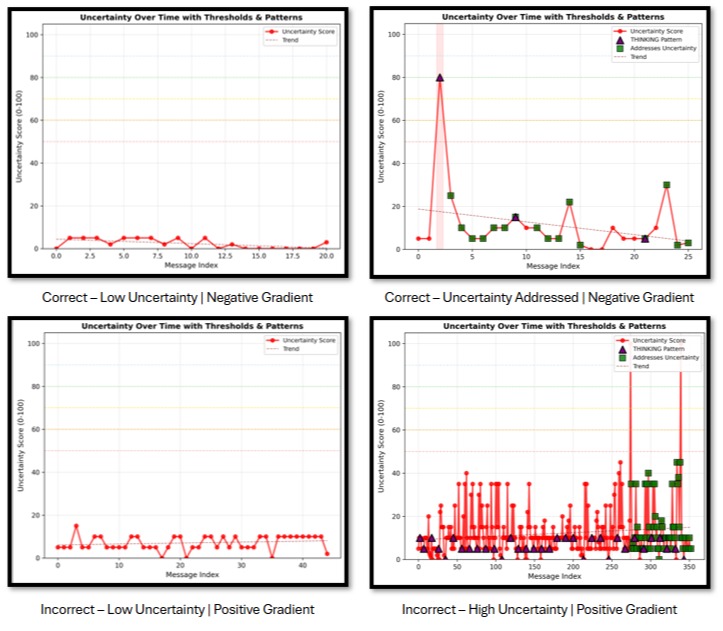}
    \caption{Direct comparison of CLIO’s internal uncertainty principles when correct and incorrect}
    \label{fig:uncertainty}
\end{figure}

Figure \ref{fig:uncertainty} depicts the dynamic nature of CLIO’s internal uncertainty measurement across four different temporally oriented views. To post-hoc analyze uncertainty within CLIO’s thinking process, we executed a series of prompts that extracted the presence, level, and prior uncertainties that had been addressed. These extracted features were subsequently utilized to construct a timestamped dataset that allowed for trend analysis. To study trends in CLIO’s uncertainties, we ran and analyzed statistical power within individual HLE questions to isolate any variability that was present, i.e. question type (multiple choice versus free response) and scope of multiple-choice question (four answers versus fifteen answers).\\

\noindent Within Figure \ref{fig:uncertainty}, the top plots capture a view of CLIO’s uncertainty in scenarios where CLIO is correct, while the bottom figures denote incorrect answers. It is immediately clear how the gradient of uncertainty with respect to time is negative for correct answers and positive for incorrect answers. The “Correct – Low Uncertainty and Negative Gradient” plot demonstrates consistently low and stable uncertainty scores with a negative gradient, indicating very little fluctuation in uncertainty and trend to a correct answer. Similarly, the “Correct – Uncertainty Addressed and Negative Gradient” shows pronounced spikes in uncertainty, which are mitigated through CLIO’s recursive thought channels as indicated by the triangles. These clean context windows enable the dynamic adaptation of thought strategies and when correct, result in a steady decline of uncertainty to converge on a result.\\

\noindent In “Incorrect – Low Uncertainty and Positive Gradient”, CLIO has little uncertainty across its entire thinking process but trends positive. Further observations of this pattern indicate that CLIO did not leverage its recursive ability to mitigate the positive shift in uncertainty. Finally, “Incorrect – High Uncertainty and Positive Gradient” shows clear signal through high volatility and continually increasing uncertainty. Analysis of these oscillation patterns indicates CLIO’s repeated attempts to address its uncertainty but fails to do so. These oscillations are clear signals for when end users must be brought in for interaction or additional review. Interestingly, CLIO still converges on an answer despite its oscillations, highlighting an important gap in existing models or systems.\\

\subsubsection{Bio/Med Full Case Study}
Additionally, to analyze the utilization of uncertainty as a consistent flag for the user, we studied observed effects within subdomains individually, i.e., immunology and biochemistry. Using a baseline of 95\% statistical power, we found that: 

\begin{table}[hbp] 
\centering  
\begin{minipage}{\textwidth}  
\begin{tabular}{lccc}  
\toprule  
\textbf{Feature} & \textbf{p-value} & \textbf{Effect size} & \textbf{Observation} \\  
\midrule  
Initial Uncertainty & 0.011437 & -0.170 & Correct samples have higher initial uncertainty \\  
Uncertainty Range & 0.040926 & -0.141 & Correct samples have higher uncertainty ranges \\  
\bottomrule  
\end{tabular}  
\end{minipage}  
\label{tab:features1}
\end{table}

\noindent However, the aggregation and dispersion of question type imbalance between multiple choice and free response questions per subdomain increases likelihood for skewed results on analysis. For example, in neuroscience, the following metrics are critical to CLIO.

\begin{table} 
\centering  
\begin{minipage}{\textwidth}  
\begin{tabularx}{\textwidth}{lccX}  
\toprule  
\textbf{Feature} & \textbf{p-value} & \textbf{Effect size} & \textbf{Observation} \\  
\midrule  
Ratio of Addressing Uncertainty & 0.000249 & -0.511 & Correct samples address uncertainty and mitigate them successfully at a higher ratio \\  
Uncertainty Slope T-Stat & 0.000396 & 0.605 & Correct samples hold a negative gradient on the path to resolving uncertainty \\  
\bottomrule  
\end{tabularx}  
\end{minipage}  
\label{tab:features2}  
\end{table}

\noindent These results demonstrate the importance of a scientist’s exposure to knowing both CLIO's trajectory and velocity of uncertainty signals. CLIO’s prioritization of developing escalation points that can be identified and relayed to a scientist enables transparency and trustworthiness. While our work demonstrates CLIO’s self-adaptation and optimization potential, this also backs the notion that successful problem solving alongside end users can be attributed to proper interjection at the right time. Incorrect or oscillatory thought patterns can be terminated directly and resumed to follow the right direction, enabling efficient human oversight.

\section{Future Work}
As we continue our development, we approach our research under the accuracy, cost, and time (ACT) framing.  

\subsection{Accuracy}
While the effect of control variables like CLIO’s temperature and depth appear to influence overall performance, further clarification and evaluation is required and is an area we look forward to continuing our research in. Here, the tradeoffs between question complexity or nuance are of key interest in determining the depth of exploration or model to utilize when instantiating CLIO. In our initial tests, we mixed reasoning and non-reasoning models in CLIO’s implementation and observed that CLIO’s uncertainty raises the relevant flags in the reasoning models. For example, we observed CLIO’s approach effectively combining GPT-4.1 and o3 to answer questions that o3 alone failed to. This observation held true in the utilization of o3 in both high- and low-reasoning effort models. In further testing, we found that this extends to a range of non-OpenAI models such as Microsoft’s Phi-4 and xAI’s Grok-4. For example, we found that our use of Phi-4 with CLIO solved new biology questions that Phi-4 alone was unable to solve. The same was true with Grok-4, where we observed CLIO's approach enabling an additional question to be answered successfully within the molecular biology domain more than half the time. Nonetheless, further assessment is required to study the effectiveness and optimization for which models to utilize for respective problem cases, as models are performant in different categories. 

\subsection{Cost}
While the recursive design of CLIO demands considerable computational resources and extended execution periods, prioritizing the pursuit of novel scientific discoveries offers a compelling trade-off between time and computational expenditure. The utility function of cost against compute is outweighed by the potential gains, both within the scientific domain but also outside. In our current work, we demonstrate CLIO’s capabilities within science, with preliminary work showing how CLIO can extend to other domain problems with similar cognitive patterns. Problems that require recursive reasoning patterns to handle nuances and solve are ripe for CLIO’s utility (i.e. legal services or financial analytics). 

\subsection{Time}
Our early work demonstrated that CLIO can successfully run and orchestrate scientific tools autonomously for over ten hours to solve complex scientific discovery problems. Further assessments are required to show the efficacy of tool integration and guidance over time from a subject matter expert. Rather than demonstrating pre- or post-hoc control of outputs, we aim to show how mid-stream steering influences scientific outcomes.

\subsection{Scaling ACT}
Across ACT, we believe that with CLIO’s consistent capacity to generate robust chains of thought, and when properly selected, downstream model training using synthetic data can be further enabled. We are actively evaluating whether CLIO's design benefits synthetic data production for model training and what additional components could be added to ensure the highest quality chain of thought selection.

\section{Conclusion}
We present CLIO, an alternative to post-training reasoning patterns prior to end-user interaction with non-reasoning models like GPT-4.1. CLIO’s adaptation at runtime leads to increased performance on difficult science problems while offering the opportunity for additional control and exposure to the end user. As we enter a landscape where long-running science agents can perform increasingly complex and difficult science functions, the ability for reasoning systems to be controlled, monitored, and corrected is essential. Replacing reasoning trained models with the ability to alter thinking patterns in real time, express the uncertainty in approaches and progress, and match the cognition approach of scientists is necessary to reach the scientific potential of a human-machine team of scientists and AI agents.\\

\noindent We present the algorithm behind how CLIO can optimize thinking patterns at runtime through orchestration to solve problems like an expert starting from a blank slate. We demonstrate and analyze CLIO’s ability to highlight its own struggles with uncertainty. To address uncertainty, CLIO intrinsically applies course correction through recursive invocation. In open scenarios, CLIO can be configured to highlight when the appropriate time is to ask for help instead of correcting itself. \\

\noindent As the capabilities and performance of reasoning-class models and systems increase, so will the expectations of scientists and end users. Scientists who bring in domain-specific tools for utilization will increasingly require nuanced, scientific decisions to be explained and exposed. Therefore, the demonstrated inference-time optimization of cognition patterns will be significant in shaping human-machine symbiosis as we explore unexplored areas in science.

\section*{End Notes}  
We thank Jason Zander and Nadia Karim for their unwavering support and guidance throughout this entire process. We extend our thanks to colleagues both inside and outside Microsoft Discovery and Quantum for sharing their insights and feedback, including Steven Truitt, Allen Stewart, Yasser Asmi, David Marvin, Harsha Nori, Scott Lundberg, Chi Chen, and Phil Waymouth. We are deeply grateful for August Laguio and his support in the graphic design of figures. All correspondence and requests for further information should be addressed to \href{mailto:newmancheng@microsoft.com}{newmancheng@microsoft.com}.  

\bibliography{sn-bibliography}% common bib file

\begin{thebibliography}{10}
\expandafter\ifx\csname url\endcsname\relax
  \def\url#1{\burl{#1}}\fi
\expandafter\ifx\csname urlprefix\endcsname\relax\def\urlprefix{URL }\fi
\providecommand{\bibinfo}[2]{#2}
\providecommand{\eprint}[2][]{\url{#2}}
\providecommand{\doi}[1]{\url{https://doi.org/#1}}
\bibcommenthead

\bibitem{GAZERANI2025149643}
\bibinfo{author}{Gazerani, P.}
\newblock \bibinfo{title}{The neuroplastic brain: current breakthroughs and
  emerging frontiers}.
\newblock \emph{\bibinfo{journal}{Brain Research}}
  \textbf{\bibinfo{volume}{1858}}, \bibinfo{pages}{149643}
  (\bibinfo{year}{2025}).
\newblock
  \urlprefix\url{https://www.sciencedirect.com/science/article/pii/S0006899325002021}.

\bibitem{PASCUALLEONE2006315}
\bibinfo{author}{Pascual-Leone, A.}
\newblock \bibinfo{title}{ in \textit{Disrupting the brain to guide plasticity
  and improve behavior}} (ed.\bibinfo{editor}{Møller, A.~R.})
  \emph{\bibinfo{booktitle}{Reprogramming of the Brain}}, Vol.
  \bibinfo{volume}{157} of \emph{\bibinfo{series}{Progress in Brain Research}}
  \bibinfo{pages}{315--404} (\bibinfo{publisher}{Elsevier},
  \bibinfo{year}{2006}).
\newblock
  \urlprefix\url{https://www.sciencedirect.com/science/article/pii/S0079612306570190}.

\bibitem{Qadir2025}
\bibinfo{author}{Qadir, H.~M.} \emph{et~al.}
\newblock \bibinfo{title}{An adaptive feedback system for the improvement of
  learners}.
\newblock \emph{\bibinfo{journal}{Scientific Reports}}
  \textbf{\bibinfo{volume}{15}}, \bibinfo{pages}{17242} (\bibinfo{year}{2025}).
\newblock \urlprefix\url{https://doi.org/10.1038/s41598-025-01429-w}.
\newblock \bibinfo{note}{Published: 2025-05-18}.

\bibitem{Don-Yehiya2025}
\bibinfo{author}{Don-Yehiya, S.} \emph{et~al.}
\newblock \bibinfo{title}{The future of open human feedback}.
\newblock \emph{\bibinfo{journal}{Nature Machine Intelligence}}
  \textbf{\bibinfo{volume}{7}}, \bibinfo{pages}{825--835}
  (\bibinfo{year}{2025}).
\newblock \urlprefix\url{https://doi.org/10.1038/s42256-025-01038-2}.

\bibitem{Kasdin2025}
\bibinfo{author}{Kasdin, J.} \emph{et~al.}
\newblock \bibinfo{title}{Natural behaviour is learned through
  dopamine-mediated reinforcement}.
\newblock \emph{\bibinfo{journal}{Nature}} \textbf{\bibinfo{volume}{641}},
  \bibinfo{pages}{699--706} (\bibinfo{year}{2025}).
\newblock \urlprefix\url{https://doi.org/10.1038/s41586-025-08729-1}.

\bibitem{Mohar2025}
\bibinfo{author}{Mohar, B.} \emph{et~al.}
\newblock \bibinfo{title}{Delta: a method for brain-wide measurement of
  synaptic protein turnover reveals localized plasticity during learning}.
\newblock \emph{\bibinfo{journal}{Nature Neuroscience}}
  \textbf{\bibinfo{volume}{28}}, \bibinfo{pages}{1089--1098}
  (\bibinfo{year}{2025}).
\newblock \urlprefix\url{https://doi.org/10.1038/s41593-025-01923-4}.

\bibitem{Clewett2019}
\bibinfo{author}{Clewett, D.}, \bibinfo{author}{DuBrow, S.} \&
  \bibinfo{author}{Davachi, L.}
\newblock \bibinfo{title}{Transcending time in the brain: How event memories
  are constructed from experience}.
\newblock \emph{\bibinfo{journal}{Hippocampus}} \textbf{\bibinfo{volume}{29}},
  \bibinfo{pages}{162--183} (\bibinfo{year}{2019}).
\newblock \bibinfo{note}{Epub 2019 Feb 7. PMID: 30734391; PMCID: PMC6629464}.

\bibitem{guo2025deepseek}
\bibinfo{author}{Guo, D.} \emph{et~al.}
\newblock \bibinfo{title}{Deepseek-r1: Incentivizing reasoning capability in
  llms via reinforcement learning}.
\newblock \emph{\bibinfo{journal}{arXiv preprint arXiv:2501.12948}}
  (\bibinfo{year}{2025}).

\bibitem{wen2025reinforcement}
\bibinfo{author}{Wen, X.} \emph{et~al.}
\newblock \bibinfo{title}{Reinforcement learning with verifiable rewards
  implicitly incentivizes correct reasoning in base llms}.
\newblock \emph{\bibinfo{journal}{arXiv preprint arXiv:2506.14245}}
  (\bibinfo{year}{2025}).

\bibitem{zhao2025learning}
\bibinfo{author}{Zhao, X.}, \bibinfo{author}{Kang, Z.}, \bibinfo{author}{Feng,
  A.}, \bibinfo{author}{Levine, S.} \& \bibinfo{author}{Song, D.}
\newblock \bibinfo{title}{Learning to reason without external rewards}.
\newblock \emph{\bibinfo{journal}{arXiv preprint arXiv:2505.19590}}
  (\bibinfo{year}{2025}).

\bibitem{su2025crossing}
\bibinfo{author}{Su, Y.} \emph{et~al.}
\newblock \bibinfo{title}{Crossing the reward bridge: Expanding rl with
  verifiable rewards across diverse domains}.
\newblock \emph{\bibinfo{journal}{arXiv preprint arXiv:2503.23829}}
  (\bibinfo{year}{2025}).

\bibitem{korbak2025chain}
\bibinfo{author}{Korbak, T.} \emph{et~al.}
\newblock \bibinfo{title}{Chain of thought monitorability: A new and fragile
  opportunity for ai safety}.
\newblock \emph{\bibinfo{journal}{arXiv preprint arXiv:2507.11473}}
  (\bibinfo{year}{2025}).

\bibitem{xu2025towards}
\bibinfo{author}{Xu, F.} \emph{et~al.}
\newblock \bibinfo{title}{Towards large reasoning models: A survey of
  reinforced reasoning with large language models}.
\newblock \emph{\bibinfo{journal}{arXiv preprint arXiv:2501.09686}}
  (\bibinfo{year}{2025}).

\bibitem{novikov2025alphaevolve}
\bibinfo{author}{Novikov, A.} \emph{et~al.}
\newblock \bibinfo{title}{Alphaevolve: A coding agent for scientific and
  algorithmic discovery}.
\newblock \emph{\bibinfo{journal}{arXiv preprint arXiv:2506.13131}}
  (\bibinfo{year}{2025}).

\bibitem{shojaee2025illusion}
\bibinfo{author}{Shojaee, P.} \emph{et~al.}
\newblock \bibinfo{title}{The illusion of thinking: Understanding the strengths
  and limitations of reasoning models via the lens of problem complexity}.
\newblock \emph{\bibinfo{journal}{arXiv preprint arXiv:2506.06941}}
  (\bibinfo{year}{2025}).

\bibitem{SLADKY2024223}
\bibinfo{author}{Sladky, R.}, \bibinfo{author}{Kargl, D.},
  \bibinfo{author}{Haubensak, W.} \& \bibinfo{author}{Lamm, C.}
\newblock \bibinfo{title}{An active inference perspective for the amygdala
  complex}.
\newblock \emph{\bibinfo{journal}{Trends in Cognitive Sciences}}
  \textbf{\bibinfo{volume}{28}}, \bibinfo{pages}{223--236}
  (\bibinfo{year}{2024}).
\newblock
  \urlprefix\url{https://www.sciencedirect.com/science/article/pii/S1364661323002838}.

\bibitem{SHEA2014186}
\bibinfo{author}{Shea, N.} \emph{et~al.}
\newblock \bibinfo{title}{Supra-personal cognitive control and metacognition}.
\newblock \emph{\bibinfo{journal}{Trends in Cognitive Sciences}}
  \textbf{\bibinfo{volume}{18}}, \bibinfo{pages}{186--193}
  (\bibinfo{year}{2014}).
\newblock
  \urlprefix\url{https://www.sciencedirect.com/science/article/pii/S1364661314000230}.

\bibitem{doi:10.1126/sciadv.adn5290}
\bibinfo{author}{Doshi, A.~R.} \& \bibinfo{author}{Hauser, O.~P.}
\newblock \bibinfo{title}{Generative ai enhances individual creativity but
  reduces the collective diversity of novel content}.
\newblock \emph{\bibinfo{journal}{Science Advances}}
  \textbf{\bibinfo{volume}{10}}, \bibinfo{pages}{eadn5290}
  (\bibinfo{year}{2024}).
\newblock
  \urlprefix\url{https://www.science.org/doi/abs/10.1126/sciadv.adn5290}.

\bibitem{chowdhery2023palm}
\bibinfo{author}{Chowdhery, A.} \emph{et~al.}
\newblock \bibinfo{title}{Palm: Scaling language modeling with pathways}.
\newblock \emph{\bibinfo{journal}{Journal of Machine Learning Research}}
  \textbf{\bibinfo{volume}{24}}, \bibinfo{pages}{1--113}
  (\bibinfo{year}{2023}).

\bibitem{Silver2016}
\bibinfo{author}{Silver, D.} \emph{et~al.}
\newblock \bibinfo{title}{Mastering the game of go with deep neural networks
  and tree search}.
\newblock \emph{\bibinfo{journal}{Nature}} \textbf{\bibinfo{volume}{529}},
  \bibinfo{pages}{484--489} (\bibinfo{year}{2016}).
\newblock \urlprefix\url{https://doi.org/10.1038/nature16961}.

\bibitem{gottweis2025towards}
\bibinfo{author}{Gottweis, J.} \emph{et~al.}
\newblock \bibinfo{title}{Towards an ai co-scientist}.
\newblock \emph{\bibinfo{journal}{arXiv preprint arXiv:2502.18864}}
  (\bibinfo{year}{2025}).

\bibitem{lu2024ai}
\bibinfo{author}{Lu, C.} \emph{et~al.}
\newblock \bibinfo{title}{The ai scientist: Towards fully automated open-ended
  scientific discovery}.
\newblock \emph{\bibinfo{journal}{arXiv preprint arXiv:2408.06292}}
  (\bibinfo{year}{2024}).

\bibitem{wang2025txgemma}
\bibinfo{author}{Wang, E.} \emph{et~al.}
\newblock \bibinfo{title}{Txgemma: Efficient and agentic llms for
  therapeutics}.
\newblock \emph{\bibinfo{journal}{arXiv preprint arXiv:2504.06196}}
  (\bibinfo{year}{2025}).

\bibitem{Huang2025.05.30.656746}
\bibinfo{author}{Huang, K.} \emph{et~al.}
\newblock \bibinfo{title}{Biomni: A general-purpose biomedical ai agent}.
\newblock \emph{\bibinfo{journal}{bioRxiv}}  (\bibinfo{year}{2025}).
\newblock
  \urlprefix\url{https://www.biorxiv.org/content/early/2025/06/02/2025.05.30.656746}.

\bibitem{ghareeb2025robin}
\bibinfo{author}{Ghareeb, A.~E.} \emph{et~al.}
\newblock \bibinfo{title}{Robin: A multi-agent system for automating scientific
  discovery}.
\newblock \emph{\bibinfo{journal}{arXiv preprint arXiv:2505.13400}}
  (\bibinfo{year}{2025}).

\bibitem{Delgado-Licona2025}
\bibinfo{author}{Delgado-Licona, F.} \emph{et~al.}
\newblock \bibinfo{title}{Flow-driven data intensification to accelerate
  autonomous inorganic materials discovery}.
\newblock \emph{\bibinfo{journal}{Nature Chemical Engineering}}
  \textbf{\bibinfo{volume}{2}}, \bibinfo{pages}{436--446}
  (\bibinfo{year}{2025}).
\newblock \urlprefix\url{https://doi.org/10.1038/s44286-025-00249-z}.

\bibitem{gridach2025agentic}
\bibinfo{author}{Gridach, M.}, \bibinfo{author}{Nanavati, J.},
  \bibinfo{author}{Abidine, K. Z.~E.}, \bibinfo{author}{Mendes, L.} \&
  \bibinfo{author}{Mack, C.}
\newblock \bibinfo{title}{Agentic ai for scientific discovery: A survey of
  progress, challenges, and future directions}.
\newblock \emph{\bibinfo{journal}{arXiv preprint arXiv:2503.08979}}
  (\bibinfo{year}{2025}).

\bibitem{Bieberich2012}
\bibinfo{author}{Bieberich, E.}
\newblock \bibinfo{title}{Introduction to the fractality principle of
  consciousness and the sentyon postulate}.
\newblock \emph{\bibinfo{journal}{Cognitive Computation}}
  \textbf{\bibinfo{volume}{4}}, \bibinfo{pages}{13--28} (\bibinfo{year}{2012}).
\newblock \urlprefix\url{https://doi.org/10.1007/s12559-011-9104-5}.

\bibitem{ferrigno2020recursive}
\bibinfo{author}{Ferrigno, S.}, \bibinfo{author}{Cheyette, S.~J.},
  \bibinfo{author}{Piantadosi, S.~T.} \& \bibinfo{author}{Cantlon, J.~F.}
\newblock \bibinfo{title}{Recursive sequence generation in monkeys, children,
  us adults, and native amazonians}.
\newblock \emph{\bibinfo{journal}{Science Advances}}
  \textbf{\bibinfo{volume}{6}}, \bibinfo{pages}{eaaz1002}
  (\bibinfo{year}{2020}).

\bibitem{wei2022chain}
\bibinfo{author}{Wei, J.} \emph{et~al.}
\newblock \bibinfo{title}{Chain-of-thought prompting elicits reasoning in large
  language models}.
\newblock \emph{\bibinfo{journal}{Advances in neural information processing
  systems}} \textbf{\bibinfo{volume}{35}}, \bibinfo{pages}{24824--24837}
  (\bibinfo{year}{2022}).

\bibitem{yao2023react}
\bibinfo{author}{Yao, S.} \emph{et~al.}
\newblock \bibinfo{title}{React: Synergizing reasoning and acting in language
  models}  (\bibinfo{year}{2023}).

\bibitem{AdahMiller2025}
\bibinfo{author}{Miller, E.~A.}, \bibinfo{author}{Li, T.},
  \bibinfo{author}{Chen, I.-C.}, \bibinfo{author}{Krajcik, J.} \&
  \bibinfo{author}{Kelly, S.~C.}
\newblock \bibinfo{title}{Designing for and investigating elementary
  students’ cognitive flexibility, science, and literacy achievement in
  project-based science learning}.
\newblock \emph{\bibinfo{journal}{Disciplinary and Interdisciplinary Science
  Education Research}} \textbf{\bibinfo{volume}{7}}, \bibinfo{pages}{13}
  (\bibinfo{year}{2025}).
\newblock \urlprefix\url{https://doi.org/10.1186/s43031-025-00131-1}.

\bibitem{liu2023lost}
\bibinfo{author}{Liu, N.~F.} \emph{et~al.}
\newblock \bibinfo{title}{Lost in the middle: How language models use long
  contexts}.
\newblock \emph{\bibinfo{journal}{arXiv preprint arXiv:2307.03172}}
  (\bibinfo{year}{2023}).

\bibitem{hsieh2024ruler}
\bibinfo{author}{Hsieh, C.-P.} \emph{et~al.}
\newblock \bibinfo{title}{Ruler: What's the real context size of your
  long-context language models?}
\newblock \emph{\bibinfo{journal}{arXiv preprint arXiv:2404.06654}}
  (\bibinfo{year}{2024}).

\bibitem{edge2024local}
\bibinfo{author}{Edge, D.} \emph{et~al.}
\newblock \bibinfo{title}{From local to global: A graph rag approach to
  query-focused summarization}.
\newblock \emph{\bibinfo{journal}{arXiv preprint arXiv:2404.16130}}
  (\bibinfo{year}{2024}).

\bibitem{whiting2024driftsearch}
\bibinfo{author}{Whiting, J.} \emph{et~al.}
\newblock \bibinfo{title}{Introducing drift search: Combining global and local
  search methods to improve quality and efficiency}.
\newblock
  \bibinfo{howpublished}{\url{https://www.microsoft.com/en-us/research/blog/introducing-drift-search-combining-global-and-local-search-methods-to-improve-quality-and-efficiency/}}
  (\bibinfo{year}{2024}).
\newblock \bibinfo{note}{Microsoft Research Blog}.

\end{thebibliography}

\end{document}